# Accurate Reconstruction of Gas Turbine Blade Geometry Using 3D/2D Rigid Registration and CT View Optimization

Hristo Valtchanov[1], Nicolas Piché[2], Vladimir Brailovski[3], Justin Byers[4], Catherine Désrosiers[1,2], François Guibault[1]

[1]École Polytechnique de Montréal, Montréal, QC, Canada

[2]Object Research Systems Inc., Montréal, QC, Canada

[3]École de technologie supérieure, Montréal, QC, Canada

[4]Pratt & Whitney Canada, Montréal, QC, Canada

**Abstract**

Non-destructive X-ray and computed tomography (CT) testing are essential for ensuring the dimensional accuracy of manufactured components with complex internal structures such as the cooling channels in gas turbine blades, which directly impact thermal performance and service life. This study presents a multi-part 3D-2D rigid registration approach for aligning CAD models to X-ray projections as an alternative to CT reconstruction for part inspection and measurement. A greedy registration algorithm is employed, sequentially aligning the blade's exterior before registering internal components to maximize mutual information between simulated and scanned X-ray images. This stepwise approach significantly reduces problem complexity and improves alignment accuracy. View-angle optimization is performed using a greedy method, iteratively selecting angles to minimize dimensional deviation in measurements. Results indicate that a small number of oblique views provide the best accuracy, but a broad range of angles can yield acceptable results. The method achieves sub-pixel registration accuracy, with errors reduced to less than one-fifth of the (magnified) pixel pitch. Image noise and defects influence registration precision, but working in the projection space mitigates these effects compared to CT reconstruction. These findings indicate that the accuracy of the registration process is sensitive to X-ray image noise and defects but appropriate view optimization can mitigate these effects so that acceptable sub-pixel accuracy is achieved.

## 1 Introduction

3D-2D image registration aligns a three-dimensional object, represented as a mesh or volume, with two-dimensional projection images. It has applications in robotics, computer vision, image-guided medical interventions, and industrial inspection. In medicine, it enables preoperative 3D models to be aligned with intraoperative 2D images for surgical guidance, radiotherapy, and diagnosis [1–4]. Industrial applications include part inspection, metrology, and non-destructive testing (NDT) [5].

This work focuses on registering multiple rigid 3D CAD components with X-ray projections as a surrogate for CT reconstruction in part inspection and measurement. The approach is particularly useful for assessing internal cavities in complex components such as gas turbine blades. 3D-2D rigid registration, also called pose estimation, determines the translation and orientation of 3D components relative to 2D images. It is also a prerequisite for elastic registration and presents distinct optimization challenges. Registration methods are generally categorized as extrinsic, which use physical markers to establish dimensional correspondence [6], or intrinsic, which rely on image data and knowledge of the imaging system. Intrinsic approaches may be feature-based [3], gradient-based [2], or intensity-based [1,7,8].

The choice of similarity metric is crucial to registration accuracy and efficiency. Mutual information (MI) is widely used for multimodal alignment [1,9], whereas normalized cross-correlation (NCC) is suitable for monomodal data with linear intensity variations [6]. The sum of squared differences and the entropy of the difference image are additional options for intensity- and feature-based methods.

In medical imaging, 3D-2D rigid registration has improved alignment precision. Tang et al. [10] used fiducial markers to achieve submillimeter accuracy in radiotherapy. Markelj et al. [11] combined RANSAC with gradient matching for CT/MR-to-X-ray registration. Uneri et al. [4] found that projection separations of approximately 20° provided accurate registration in minimally invasive surgery. Wang et al. [12] introduced a point-to-plane model for accurate single-view X-ray registration. Schaffert et al. [13] extended this approach to multiview cerebral angiography, while Sun et al. [14] improved computational efficiency using perspective-projection triangular features.

In NDT, 3D-2D rigid registration aligns CAD models with X-ray projections for structural and dimensional assessment. Aouadi and Sarry [15] applied intensity-based MI registration to X-ray images and achieved submillimeter accuracy, although local minima and attenuation noise remained challenging. Mery [16] described multiview geometry for automated X-ray inspection, improving flaw detection in castings while highlighting calibration and geometric-distortion issues. Bussy et al. [17] aligned CAD models with 2D projections for sparse-view CT, improving reconstruction efficiency but encountering ambiguities in symmetric features. Tan et al. [18] matched geometric features extracted from X-ray images with projected CAD-mesh contours, thereby avoiding CT artifacts and enabling accurate inspection of metal parts, although highly attenuating materials remained challenging.

## 1.1 Multi-part registration

Multipart 3D-2D rigid registration addresses the alignment of segmented or articulated 3D models with 2D projection images. It is a constrained subset of elastic registration and must account for occlusion, clutter, and components that may not be visible in every view. Yonemoto et al. [19] introduced deformable super quadrics in an analysis-by-synthesis framework to address self-occlusion using multiple calibrated cameras and iterative refinement. Litany et al. [20] extended iterative closest point algorithms with regularization to assemble fragmented objects in the presence of missing parts or clutter. More recent work has emphasized accuracy and robustness. Wang et al. [21] used a point-to-plane correspondence model that updates 3D-2D correspondences component by component, achieving submillimeter accuracy in single-view X-ray registration. Markelj et al. [11] combined gradient-based registration with RANSAC to handle noisy data and a limited number of X-ray views.

## 1.2 Effects of Image Quality and View Angles on Registration Accuracy

The accuracy and robustness of 3D-2D registration depend strongly on image quality, noise, and the number and angular separation of projection views. Tomaževič et al. [22] showed that additional X-ray views improve robustness and reduce failure rates, but do not necessarily improve accuracy; a small to moderate number of oblique views was often optimal. Uneri et al. [4] reported that two views separated by only 10°–20° could achieve submillimeter accuracy in minimally invasive surgery and that larger separations generally improved triangulation and localization. Image preprocessing can also be important. Kim et al. [23] found that histogram equalization widened the capture range of intensity-based registration and reduced errors to 0.12° in rotation and 0.47 mm in translation. Conversely, noise and limited-angle acquisition can reduce accuracy and repeatability [24]. Zhang [24] proposed an unsupervised deep-learning framework to improve correspondence in sparse-view or low-quality imaging scenarios.

This study addresses the alignment of 3D CAD models, represented by triangle meshes, with acquired 2D X-ray projections. The following sections describe the method and evaluate rigid-registration accuracy, view-angle selection, and sensitivity to image noise.

# 2 Methods

X-ray projections (radiograms) are acquired using an FF-35 system and a linear diode array (LDA) system. The CAD model is divided into components that are registered individually with the corresponding 2D projections. The study has three objectives: to develop an automated method for estimating the pose of segmented CAD components with accuracy comparable to CT-based measurement; to determine view angles that improve accuracy and reliability; and to evaluate whether the method can achieve subpixel accuracy relative to a nominal manufacturing target of 0.2 pixels.

## 2.1 Multipart Registration Procedure

The multipart registration procedure begins by segmenting the CAD geometry into external and internal components whose relative poses may vary. Each component can undergo an independent rigid translation and rotation. Registration proceeds greedily: at each step, partial simulated radiograms are computed from the component being optimized together with all previously registered components, while components not yet registered are excluded.

Registration begins with the outermost component and proceeds inward. For each component, translation and rotation are obtained by maximizing the mutual information between the simulated and acquired radiograms over the selected views. The registration problem is formulated as follows:

$$\arg\ max_{T_i} \sum_{v}^{V} MI\big(I_{sim,v}(T_i), I_{obs,v}\big)\ \forall\ v \in V = \{v_1, v_2, \dots, v_V\}$$

$$x_i' = R(\epsilon_i, \bar{x}_0)x_i + \delta_i$$

Here, $T_i$, represents the transformation parameters (translation, $\delta_i$, and rotation about the centroid of the 0$^{th}$ component, $R$) of the $i$-th component currently being registered, $I_{sim,j}$ is the simulated X-ray image at view $j$, $I_{obs,j}$ is the observed X-ray image, and $j$ denotes the set of view positions. $MI$ denotes the mutual information.

After the pose of a component is estimated, its position is updated and held fixed while the next component is registered. At each step, the partial simulated radiogram includes the current component and all previously registered components but excludes components not yet registered. A depth map is computed for each component by casting rays from the source through the centers of the detector pixels. The component depth maps are then combined to produce a composite partial radiogram.

## 2.2 Simulated X-ray Projections and Objective Functions

To compute simulated X-ray projections efficiently, depth-map operations are used instead of explicit geometric Boolean operations among all components. Rays are cast from the X-ray source through each detector pixel and intersected with the component meshes. The resulting entry and exit depths are combined so that the path length through enclosed voids is subtracted from the total path length through solid material. Figure 1 illustrates this procedure for a representative geometry. The simulator was implemented either with ray casting in Open3D or with rasterization and projection transformations in PyTorch3D.
The composite depth map is transformed using an exponential attenuation model to create a simulated radiogram, and noise is optionally applied to account for real-world variability. Mutual information, $MI$, is then computed for each projection view v:

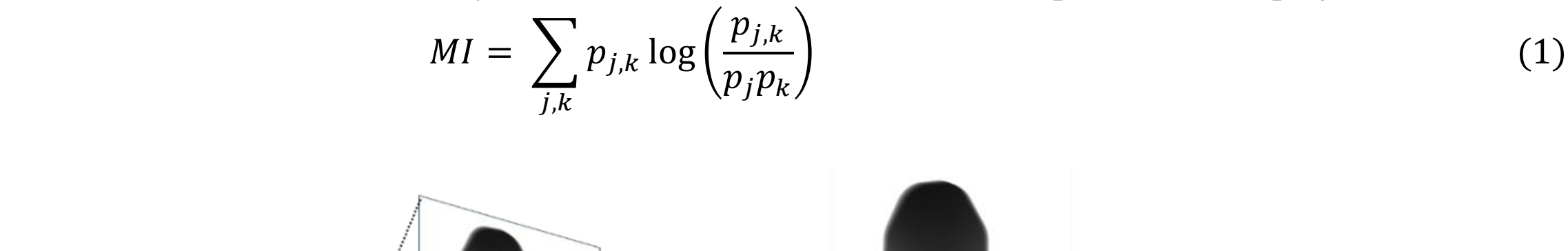

$$MI = \sum_{j,k} p_{j,k} \log\left(\frac{p_{j,k}}{p_j p_k}\right) \quad (1)$$

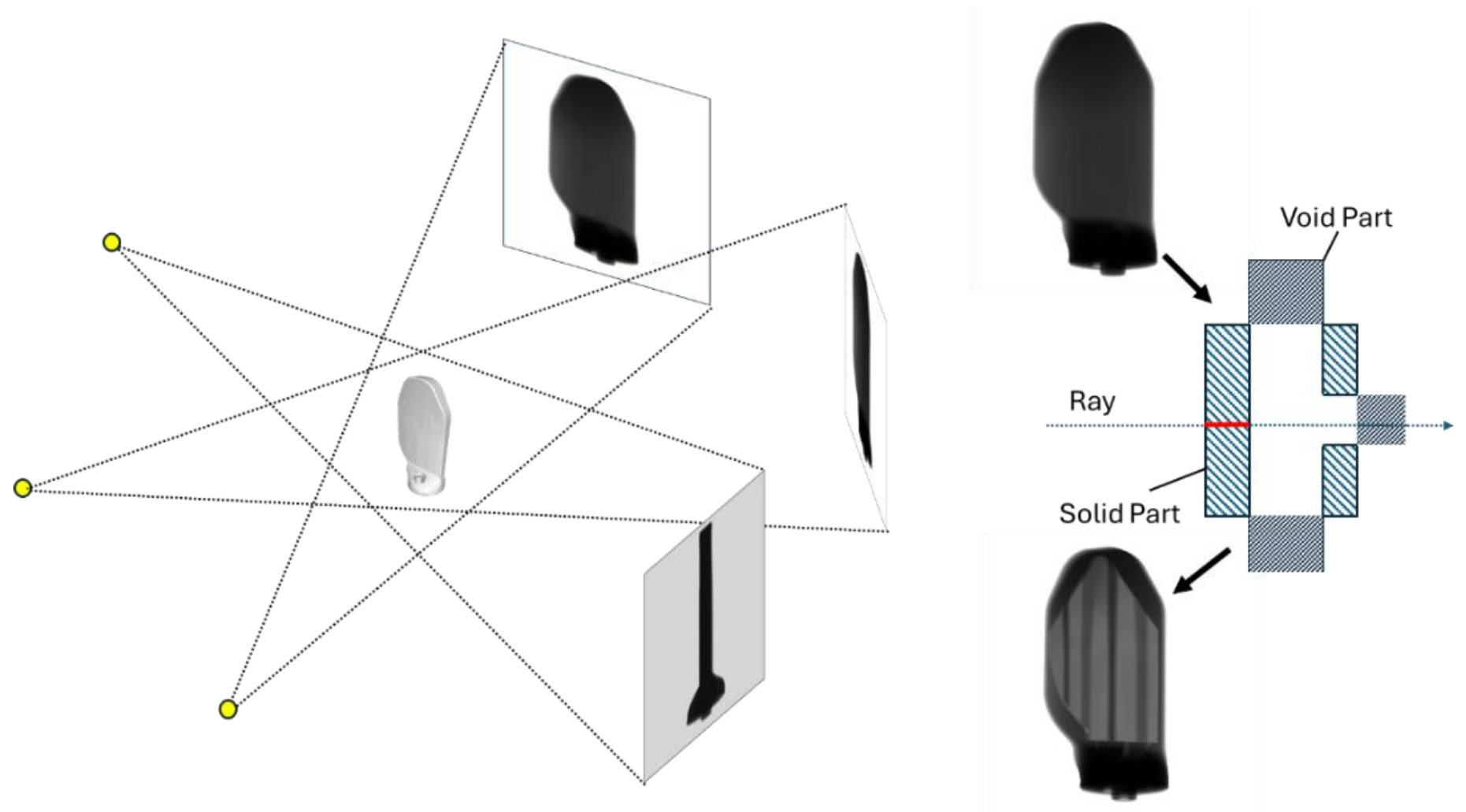

*Figure 1: Schematic of simulated-radiogram generation. Rays originate from a point source, form a cone beam, and pass through the center of each detector pixel. Internal voids are handled by combining overlapping solid and void depth maps and calculating the path length through solid material while excluding the path through voids (red).*

Here, $p_{j,k}$ is the joint probability distribution (2D histogram) of the source and target images, $p_j$ and $p_k$, are their respective individual probability distributions, and j and k denote their respective set of histogram bins. The mutual information is bounded from above by the minimum of the Shannon entropy, H, of either image, which is used as a normalization metric

$$NMI = \frac{MI}{\min\left(H(p_j), H(p_k)\right)}, H(J) = -\sum_{j\in J} p_j \log(p_j) \quad (2)$$

## 2.3 Optimization approach

Several optimization algorithms were evaluated in terms of accuracy, robustness, and computational cost, including simulated annealing, mesh adaptive direct search (MADS), BFGS, Nelder-Mead simplex, COBYLA, sequential least-squares programming (SLSQP), and Powell's conjugate-direction method. In this application, gradient-based methods such as BFGS and gradient descent either converged to local optima or were prohibitively slow. Simulated annealing, MADS, SLSQP, and simplex methods also failed to converge reliably because the objective landscape contains numerous local extrema, as illustrated in Figure 2.

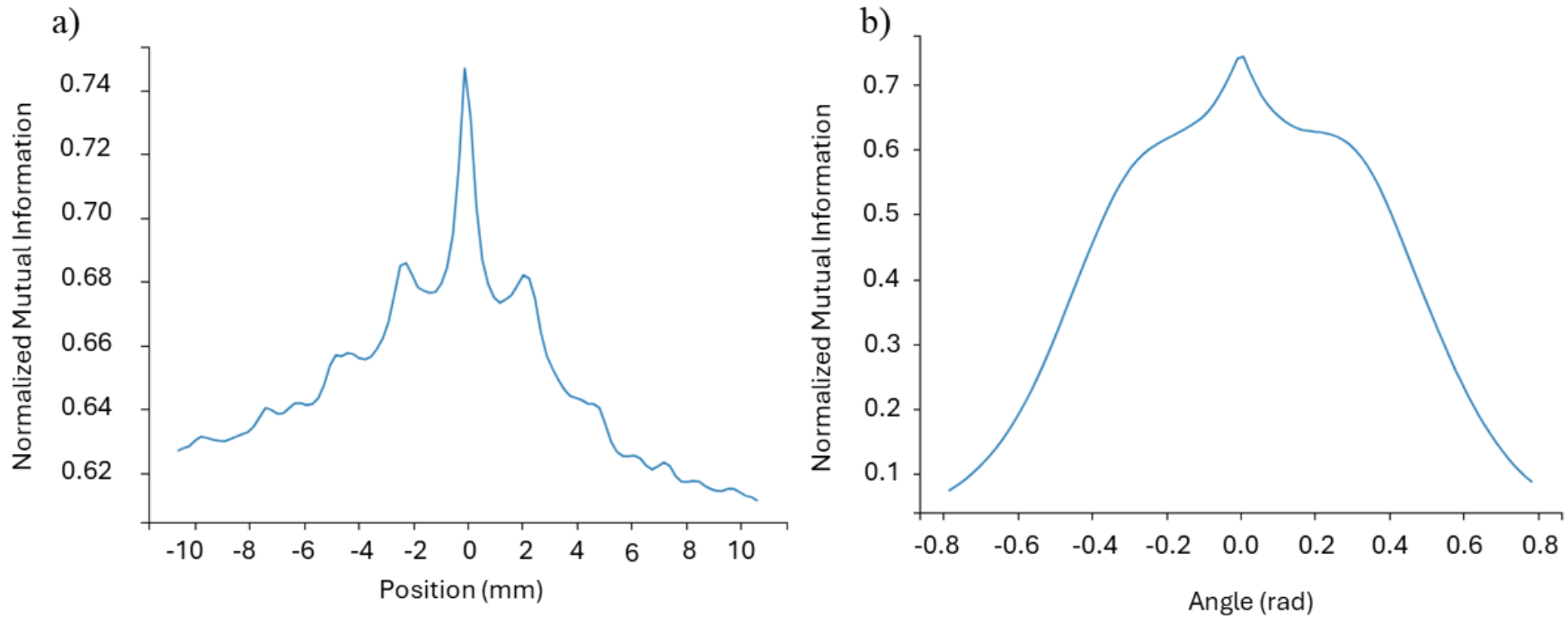

*Figure 2: Normalized mutual information as a function of registration parameters: (a) positional deviation from the reference pose along one dimension and (b) rotational deviation about the vertical axis.*

Powell's conjugate-direction method was selected because it does not require gradient evaluations and performed well on the nonconvex objective function. Its efficiency nevertheless depends strongly on the starting point and initial search directions; poorly chosen directions can lead to unnecessarily broad searches and long computation times as dimensionality increases. Repeated line searches along conjugate directions provide systematic exploration of the parameter space, but, as with other local derivative-free methods, convergence to the global optimum is not guaranteed.

Powell's conjugate-direction method was modified to reduce the number of iterations while maintaining accuracy. The modifications were as follows:

- Initial line-search directions are estimated from the gradient at the starting point.
- Principal component analysis (PCA) is applied to objective-function values obtained from random perturbations around the initial point to identify dominant directions of influence. Alternatively, a Hessian matrix is estimated stochastically to approximate dependencies among parameters.
- Parameters with high mutual covariance are grouped according to their joint influence on the objective function. The partial gradient associated with each group is then used as a line-search direction.
- After each iteration, the partial gradient is recalculated for the new point, and the $d-1$ previously used directions (where $d$ is the number of grouped parameters) are subtracted from the current search direction.
- Parameter groups are redefined after a specified number of iterations to account for changes in parameter covariance.

By reducing the number of search directions, this strategy decreases the number of iterations required for high-dimensional optimization while retaining the robustness of Powell's conjugate-direction method.

## 2.4 View-angle optimization and evaluation of error

View angles are optimized greedily by adding and refining one angle at a time. At each objective-function evaluation, eight registrations are performed from initial core poses subjected to random perturbations of predefined amplitude.

Part thickness measurements are calculated by determining the distance ($\delta$) between the first intersections of the inner and outer meshes at predefined points fixed relative to the exterior of the mesh. The deviations of these measurements from a known ground truth are computed to evaluate the accuracy of the registration. The optimization objective function incorporates both the mean accuracy of measurements and the probability of measurements falling below a specified threshold:

$$\arg\max_{v}\left(P\left(\left(\delta_p - \delta_{p,GT}\right)(V_{1\ldots v}) < \epsilon_{target}\right)\right)$$

Here, $\delta_p$ and $\delta_{p,GT}$ are the measured and ground-truth thicknesses at point p. The single parameter, v, is solved using the Brent algorithm in SciPy, with bounds of $[0,2\pi]$.

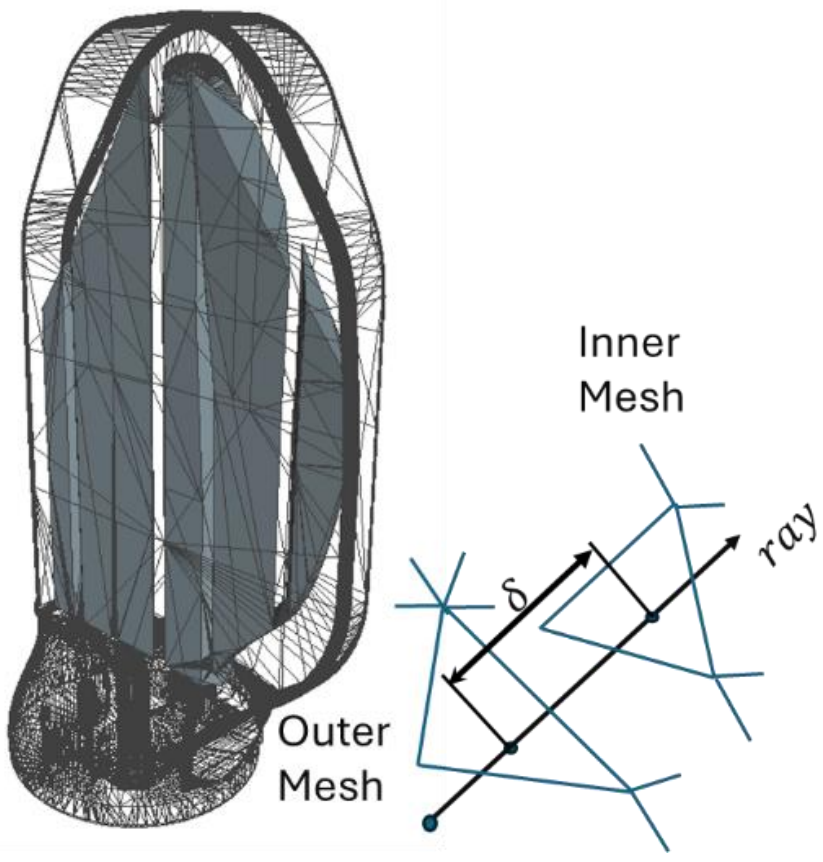

*Figure 3: Schematic of the thickness measurement. A ray is cast from a known point fixed relative to the exterior mesh, and thickness is measured between its first intersections with the exterior and interior meshes.*

Finally, zero-mean Gaussian noise is added independently to each pixel. The resulting pixel intensity is given by:

$$I'(j) = I(j) + \mathcal{N}(0,\sigma)$$

Here, $\mathcal{N}(\mu,\sigma)$ denotes a normal distribution, and $I(j)$ is the pixel intensity. The noise is zero-mean, and a uniform standard deviation of 0.025 is used for all pixels. Images are renormalized so that pixel intensity varies between 0 and 1 after noise addition. This level of noise reproduces the normalized mutual information obtained for actual CT images, which varies between 0.5 and 0.6, depending on the image quality.

## 2.5 Test cases and measurements for validation

The algorithm is evaluated in three computational studies: registration of two Pratt & Whitney turbine-blade designs (PWA and PWC) and view-angle optimization with and without noise. Radiograms of the PWC blade and the representative blade in Figure 1 were acquired with a high-resolution FF-35 industrial CT system, whereas the PWA radiograms were acquired with an LDA CT system. Synthetic reference data were used for view-angle optimization because CT reconstruction is less precise than the intended metrology system and because simulated target images can be generated efficiently for arbitrary views. For the experimental PWC cases, reference thicknesses were obtained by equivalent ray intersections with segmentation surfaces derived from CT reconstructions.

## 3 Results

The registration method was evaluated on two gas turbine blades, with accuracy assessed against measurements derived from CT reconstruction. Figure 4 illustrates the registration process.

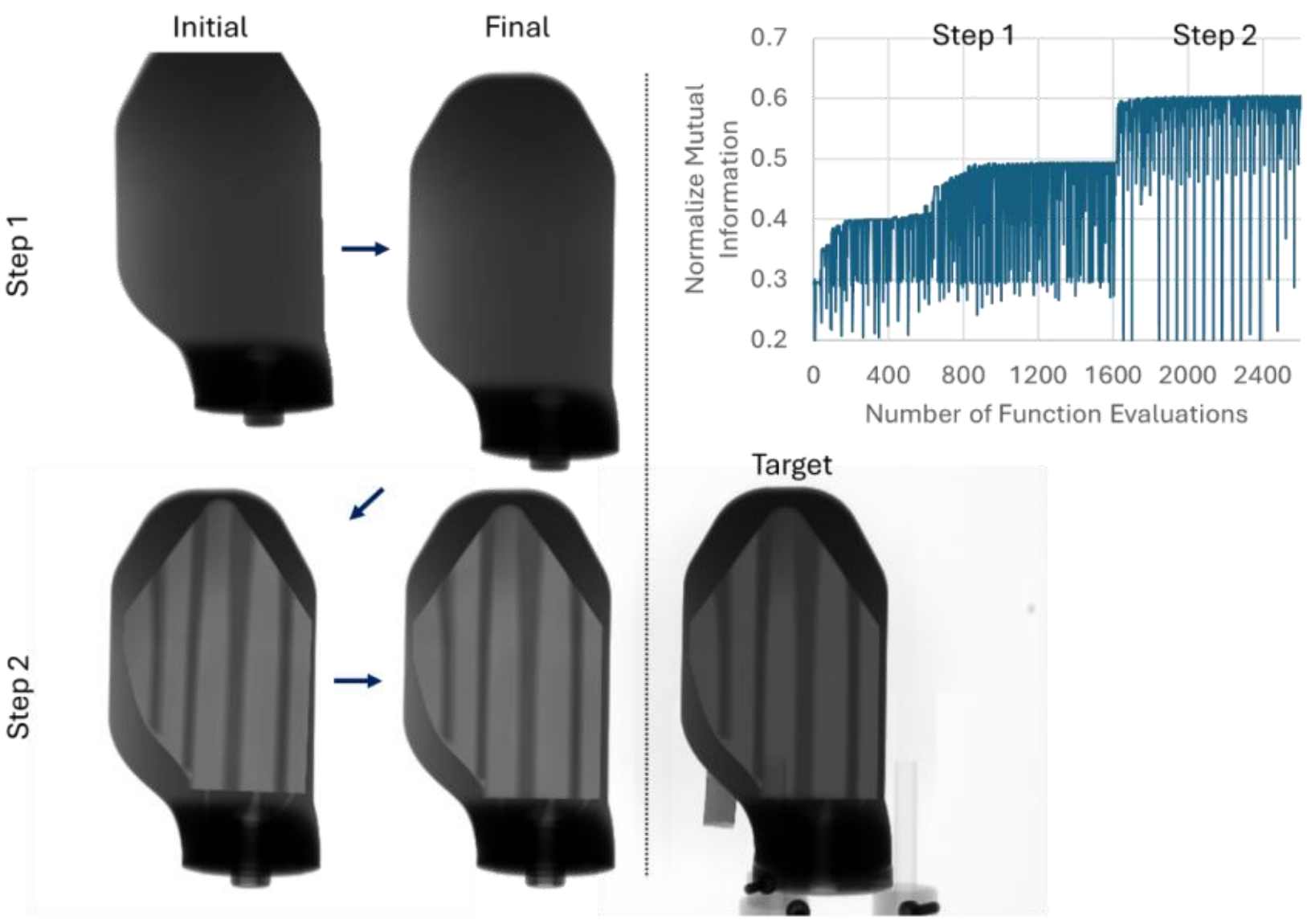

Figure 4: Greedy registration of a two-component mesh with an internal void. Top: target and simulated radiograms with the normalized-mutual-information convergence history. Bottom: sequential alignment of the exterior and internal components.

Partial radiograms of the exterior components were simulated and compared with images acquired by the FF-35 CT system. In the experiments, each rigid component produced a distinct maximum of mutual information near its best-fit alignment. The components could therefore be aligned sequentially, and the mutual information generally increased as each component was added. This intensity-based approach was more effective than direct edge alignment for the tested data.

Table 1 compares registration-derived thicknesses with the CT-derived reference measurements. The root-mean-square errors were 0.42 pixels for the PWA blade and 0.36 pixels for the PWC blade, demonstrating subpixel agreement.

Table 1: Comparison of thickness measurements obtained from CT reconstruction and registration. Sampling locations form a uniform grid on the exterior blade surface, and thicknesses are computed as shown in Figure 3. Measurements and root-mean-square errors are normalized by the magnified detector-pixel size.

a) PWA blade – RMS error - 0.423333

| Meas. Pt. | CT Reconstruction | Registration |
|---|---|---|
| 1 | 7.53 | 7.53 |
| 2 | 7.53 | 8.78 |
| 3 | 8.78 | 8.78 |
| 4 | 7.53 | 7.53 |
| 5 | 5.02 | 5.02 |
| 6 | 3.76 | 3.76 |
| 7 | 6.27 | 6.27 |
| 8 | 5.02 | 5.02 |
| 9 | 5.02 | 5.02 |
| 10 | 5.02 | 5.02 |

b) PWC Blade – RMS error 0.36

| Meas. Pt. | CT Reconstruction | Registration |
|---|---|---|
| 1 | 7.20 | 6.77 |
| 2 | 7.62 | 7.20 |
| 3 | 6.56 | 5.50 |
| 4 | 8.04 | 7.20 |
| 5 | 5.50 | 5.29 |
| 6 | 4.66 | 5.08 |
| 7 | 4.66 | 4.87 |
| 8 | 6.56 | 6.35 |
| 9 | 7.62 | 7.20 |
| 10 | 8.68 | 6.77 |

The segmentation surfaces and registration results showed good overall agreement. Some local discrepancies may reflect elastic deformation, which rigid registration cannot capture. CT reconstruction was affected by noise caused by the high X-ray attenuation and scattering characteristic of gas turbine blades. Despite poorly defined edges in some CT reconstructions, the projection-based registration remained stable. In several cases, registration also succeeded when CT reconstruction required extensive manual intervention or was infeasible.
Figure 5 shows the accuracy obtained when selecting the second view with a stochastic grid search, whereas the more efficient Brent method was used during view-angle optimization. RMS error peaked near angular separations of 0, π, and 2π, whereas oblique separations near π/2 and 3π/2 were most accurate, in agreement with Tomaževič et al. [22]. Outside these unfavorable orientations, the effect of angle was modest. Figure 6 compares greedy view-angle optimization with and without noise. The noise-free simulations were approximately one order of magnitude more precise. Increasing the number of views generally improved reliability, as indicated by the smaller standard deviations, but produced little further improvement in mean accuracy after the second or third view. Appropriate view selection nevertheless reduced the effect of noise, with a minimum error of approximately 0.05 pixel. The angular trend was consistent with that observed in Figure 5.

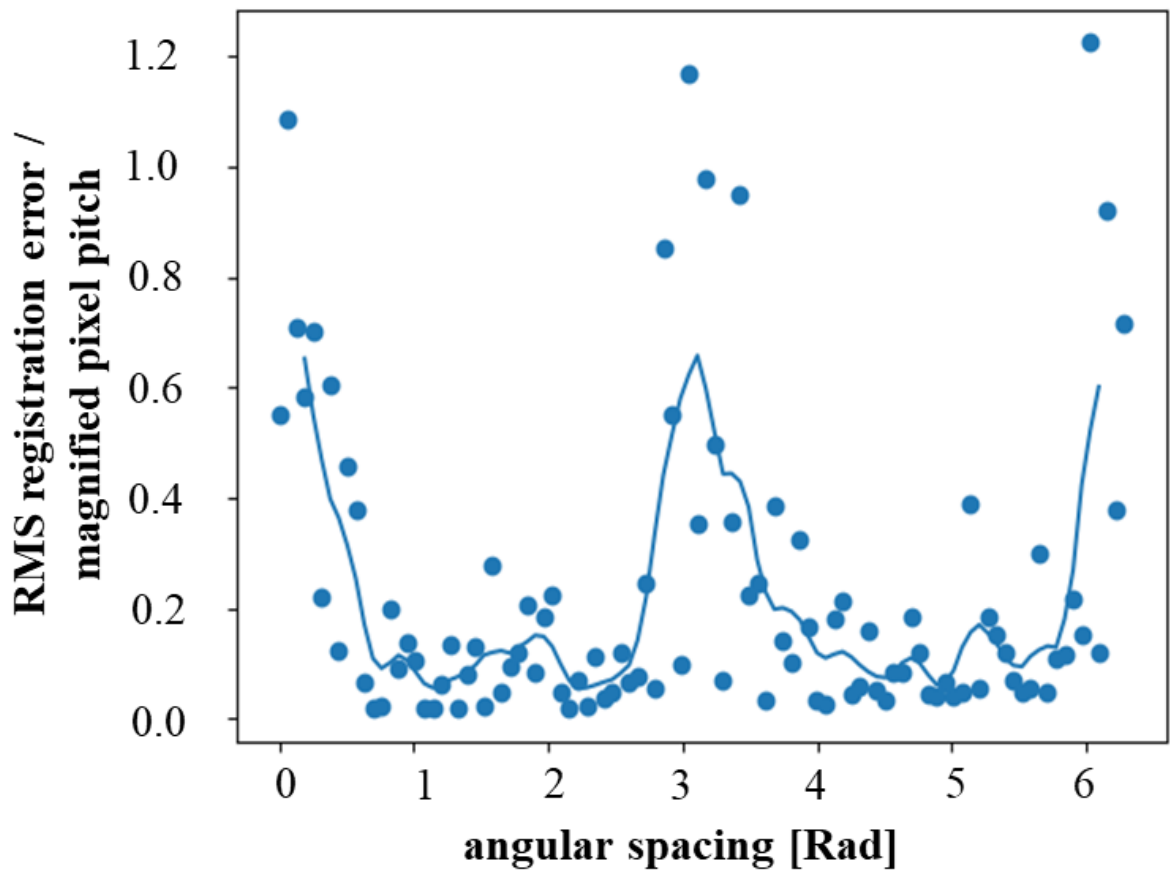

Figure 5: RMS registration error, normalized by the magnified detector-pixel size, when adding a second view at different angular separations using stochastic grid-search optimization with FF-35 radiograms.

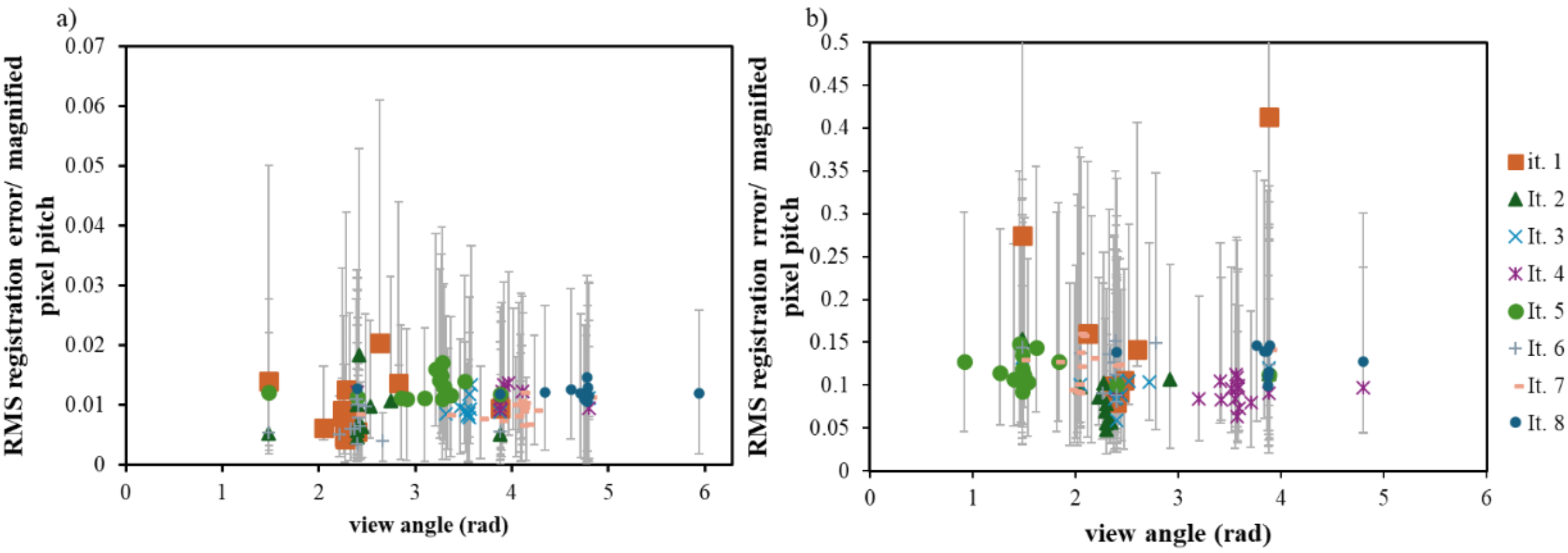

Figure 6: Greedy view-angle optimization using simulated target images (a) without noise and (b) with Gaussian noise. Each point summarizes eight registrations initialized with randomly directed pose perturbations; error bars denote standard deviations.

## 4 Discussion

The proposed method aligns components sequentially using a greedy strategy. For the tested rigid components, each partial radiogram produced a well-defined maximum of mutual information near the best-fit pose. Elastic registration introduces many additional degrees of freedom and a more complex objective landscape; performing rigid registration first is therefore expected to reduce the difficulty of the subsequent elastic-registration problem.

Subpixel accuracy was achieved with mutual information because the method uses the full intensity distribution rather than relying only on extracted edges, and constraints from multiple views can locate the optimum between detector pixels. The main sources of uncertainty were image noise, the quality of CT-derived segmentation, reference-coordinate accuracy, the deliberately coarse mesh resolution, and possible misplacement of measurement points. Elastic deformation of the manufactured part, which is not modeled by rigid registration, may also have contributed. Although many measurements agreed closely with the CT-derived reference values (Table 1), occasional larger local discrepancies were observed. Because a rigid-pose error would normally produce a more spatially coherent bias, these local errors likely arise from CT-reconstruction uncertainty, measurement-point misalignment, or unmodeled elastic deformation. With simulated radiograms as reference images, the algorithmic error was 0.05–0.1 pixels (Figure 6a).

CT reconstruction also introduced uncertainty into the reference measurements because the true surface was difficult to locate when edges were poorly defined, particularly on deep concave regions such as the pressure side of the airfoil. This issue was most evident for the PWC blade: its segmentation surface was defined manually, and the FF-35 radiograms produced a relatively noisy reconstruction. By contrast, the PWA blade was segmented in VGStudio from LDA data with less scattering, yielding more clearly defined CT surfaces. In several cases, projection-based registration succeeded even when CT reconstruction failed because of poor radiogram quality or severe scattering. This may be because relevant edges remain clearer in individual projections and because direct projection-space registration avoids the noise amplification associated with reconstruction from a limited set of views.

View angles were optimized greedily by adding and refining one angle at a time. Because the registration trials used randomly perturbed initial poses, the resulting objective values were stochastic. Noise increased the registration error by approximately one order of magnitude, but optimized view selection preserved subpixel precision. Consistent with Tomaževič et al. [22], oblique views were most effective (Figure 5). Poorly selected views reduced accuracy, whereas selecting the best view produced only modest gains once informative views were already present. Adding overlapping or redundant views could also bias the solution and slightly reduce accuracy. Reliability, measured by the standard deviation in Figure 6b, nevertheless improved as the number of views increased. Although Brent search was computationally efficient, it did not exhaustively sample the stochastic, multimodal angle space; a grid search may therefore provide a more reliable global comparison of candidate views.

## 5 Conclusion

This study demonstrated a greedy strategy for 3D-2D rigid registration in which CAD components are aligned sequentially by maximizing mutual information. The method achieved subpixel accuracy: noise-free simulations yielded errors of 0.05–0.1 pixels, while comparisons with CT-derived experimental measurements yielded root-mean-square errors of 0.36–0.42 pixels. Greedy view-angle optimization improved reliability and helped limit the effect of image noise. Adding views generally reduced variability, but provided little improvement in mean accuracy after the first few informative angles and could be detrimental when views were redundant. These results support the use of sequential rigid registration and view selection for projection-based inspection of aerospace components and as an initialization step for future elastic registration.

**Acknowledgements**

Pratt & Whitney (Raytheon Technologies) is gratefully acknowledged for providing the components and high-precision radiographic data used to validate the method. Funding was provided by PRIMA.